\documentclass[11pt,a4paper]{article}
\usepackage[hyperref]{acl2020}
\aclfinalcopy % Uncomment this line for the final submission

\usepackage[utf8]{inputenc} % allow utf-8 input
\usepackage[T1]{fontenc}    % use 8-bit T1 fonts
\usepackage{hyperref}       % hyperlinks
\usepackage{url}            % simple URL typesetting
\usepackage{booktabs}       % professional-quality tables
\usepackage{amsfonts}       % blackboard math symbols
\usepackage{nicefrac}       % compact symbols for 1/2, etc.
\usepackage{microtype}      % microtypography
\usepackage{xcolor}
\usepackage{color}
\usepackage{colortbl}
\usepackage[colorlinks]{}
\usepackage{wrapfig,lipsum,booktabs}
\usepackage{listings}
\usepackage[colorlinks]{}
\usepackage{subcaption}
\usepackage{graphicx}

\title{Athena: Constructing Dialogues Dynamically with Discourse Constraints}
\author{
	Vrindavan Harrison, Juraj Juraska, Wen Cui, Lena Reed, Kevin K. Bowden, Jiaqi Wu, \\
	\textbf{Brian Schwarzmann, Abteen Ebrahimi, Rishi Rajasekaran, Nikhil Varghese,} \\
	\textbf{Max Wechsler-Azen, Steve Whittaker, Jeffrey Flanigan}, and \textbf{Marilyn Walker} \\
	University of California, Santa Cruz \\
	Santa Cruz, CA  \\
	\texttt{\{vharriso, jjuraska, wcui7, lireeed, kkbowden, } \\
	\texttt{jwu64, brschwar, aaebrahi, rrajasek, nivarghe,}  \\ 
	\texttt{max, swhittak, jmflanigan, mawalker\}@ucsc.edu} \\
}

\begin{document}

\maketitle

\begin{abstract}
    % The abstract is submitted as plain Unicode text, with a maximum of 2000 characters.
  %Athena was developed by the NLDS Lab at UC Santa Cruz as participants in the   
  This report describes Athena, a dialogue system for spoken conversation
  on popular topics and current events. 
  We  develop a flexible topic-agnostic approach to dialogue management that dynamically configures dialogue based on general principles of entity and topic coherence.
  Athena's dialogue manager uses a contract-based method where discourse constraints are dispatched to clusters of response generators. This allows Athena to procure responses from dynamic sources, such as knowledge graph traversals and feature-based on-the-fly response retrieval methods. 
  After describing the dialogue system architecture, we perform an analysis of conversations that Athena participated in during the 2019 Alexa Prize Competition. We conclude with a report on several user studies we carried out to better understand how individual user characteristics affect system ratings.
  %, and to collect qualitative feedback on design decisions for Athena's dialogue manager.  
  
\end{abstract}

\section{Introduction}
% Davan I smoothed this here, I think what you wrote is good so I integrated these different parts (MAW Wed AM)
There has been tremendous progress over the last 10 years on  conversational agents, and a number of practical systems have been deployed. The Alexa Prize competition seeks to stimulate research and development on conversational AI in the area of open-domain topic-oriented dialogue, with recent systems contributing new ideas and methods \cite{fang2018sounding18,chen2018gunrock,bowden2019slugbot}. However the longstanding tension between hand-scripting the dialogue interaction and producing systems that scale to new domains and types of interaction still remains \cite{eric2019multiwoz,Cervoneetal19,Walkeretal07}. Dialogue systems that are trained end-to-end are not yet at a point where they can be used with real users.  The Athena system is based on a novel dialogue management architecture, which is not neural, but which aims to be scalable and dynamic, by generatively producing many different interactions for any topic using discourse constraints.

A key aspect of the Athena dialogue management architecture is the existence of multiple response generators (RGs) for each topic, which can be flexibly interleaved during a particular interaction. This emphasis on interleaving RG responses is explained and exemplified in multiple places below: here we explain why we consider it key. When constructing a dialogue system for chit-chat conversations, a commonly used approach is to construct a large and detailed dialogue plan prior to the conversation beginning. This dialogue plan can then be represented using a large static directed graph where the nodes are the system utterances and outgoing  edges are the possible user replies. During execution, a conversation is created by traveling a path through the graph. We call this the conversation flow-graph approach. Such a system can also be viewed as a finite state machine where the nodes are the states. This approach has basically not changed for over 20 years \cite{Seneff98,buntshuchetal98}, but is well-known to have severe limitations. There is virtually an unlimited number of things that the user can say next, yet the graph will always have limited out edges. Typically these conversation graphs are hand constructed which is a laborious endeavor and does not scale well. Also, this approach encourages a monolithic solution because a distributed approach requires subcomponents to pay the overhead cost of continuously orienting themselves within the graph. 

\begin{table}[ht]
%\begin{wraptable}{r}{0.4\textwidth}
%\begin{wraptable}{r}{0.65\textwidth}
% \vspace{-0.15in}
\begin{small}
%\begin{tabular}{p{0.18in} p{1.8in}}
\begin{tabular}{p{0.18in} p{2.5in}}
\textbf{RG} & \textbf{Utterance} \\ \toprule
\rowcolor[HTML]{EFEFEF}  User & that's a really interesting fact \\
DRG & I love reading about superheroes. My favorite female superhero is Black Widow. But I wanted to ask you. Are you also interested in caped crusaders? \\
\rowcolor[HTML]{EFEFEF}  User & yes i love superheroes \\
SRG & I am familiar with DC and marvel. Who is your favorite character? \\
\rowcolor[HTML]{EFEFEF}  User & most likely spider-man \\
SRG & spider-man, that's cool. Why do you love spider-man? \\
\rowcolor[HTML]{EFEFEF}  User & he's very funny and you can get around places fact \\
FRG & Do you like watching all the different marvel cinematic universe movies? \\
\rowcolor[HTML]{EFEFEF}  User & yes they are very good do you like them \\
FRG & I see. I think it would be fun to discuss all the hypothetical superpowers with my friends. What would you say is the single most awesome thing about the marvel cinematic universe? \\
\rowcolor[HTML]{EFEFEF}  User & all the crazy creative powers that stanley gave his characters \\
FRG & Yes. To me, the universe building is just really fascinating. There are not many franchises which have more than twenty different movies. \\
\rowcolor[HTML]{EFEFEF}  User & wow i didn't know that \\
CRG & I think it's cool that marvel comics has been around for so long. Marvel published a Generic Comic Book, in order to trademark the names "Super-Hero" and "Super-Villain" 
\\ \bottomrule
\end{tabular}
\caption{Example Athena conversations illustrating interleaving RG behavior among the Dialogue manager (DRG), SlugBot (SRG), Flow-RG (FRG) and Centering RG (CRG).}
\label{table:interleaving-conversation}
\end{small}
%\vspace{-0.3in}
\end{table}

In contrast, the use of interleaving RGs demonstrates Athena's dialogue manager's ability to grow conversations through incorporation of responses from new and independent sources. This is exemplified in Table~\ref{table:interleaving-conversation},\footnote{In accordance with the Alexa Prize Challenge rules, we do not share any conversations between Athena and real users.  Table~\ref{table:interleaving-conversation} and all other example conversations contained within
this document were conducted by members of our team to illustrate the functional capabilities of our socialbot.} 
where each response is labelled with the  RG that provided it, in this case four different RGs, each of which we will  explain in more detail below. By eschewing a graph based representation of dialogue state, Athena’s dialogue manager is flexible enough to use response sources that were not planned out prior to the conversation starting, and that do not need to follow rigid guidelines. We believe the dialogue management approach we developed for Athena promises future growth, and to scale to deeper and richer conversations, while at the same time encompassing a greater number of conversation topic domains.  

 %\vspace{-.2in}

\subsection{Design Philosophy and Goals}
%\noindent{\bf Design Philosophy and Goals.}
The Athena system was designed from scratch  using the Cobot Toolkit, drawing lessons from our previous competition systems  \cite{khatri2018advancing}. We developed Athena with a design philosophy that reflected  several dialogue system behavioral and design goals. Specifically, our aims are for Athena to be responsive, dynamic, modular, and convey a consistent  persona.

\noindent
\textbf{Responsive.} Athena should be responsive to user-directed conversation management cues, i.e., action directives, navigation commands, requests, etc. In particular, use of \textit{yes-no-questions} are to be avoided as a mechanism for navigating conversation branches. 

\noindent
\textbf{Dynamic.} Athena uses dynamic conversation plans and policies that move away from the handcrafted conversation flows that have become mainstays of socialbots in previous Alexa Prize competitions. This is achieved by procuring responses from dynamic sources, such as knowledge-graph traversals and feature-based on-the-fly response retrieval methods. Although we are not able to completely avoid crafted conversation flows, we limit their length to a maximum of 2 or 3 turn conversation segments. Athena is explicitly designed to allow multiple RGs on the same topic to dynamically be interleaved and thus contribute to subdialogues on a single topic. 
In effect, this causes RGs 
%(a.k.a. miniskills, scripted conversation flows, deterministic state machines, ... etc) 
to cede control, thereby opening the conversation up to dynamic response sources. 

\noindent
\textbf{Modular.} Athena's components are modular with clearly defined inputs and outputs at each stage of the system. This allows for easy updates and replacement of system components.   

\noindent\textbf{Persona.} Athena's  conversation style is not intended to mimic a human persona. Rather, our goal was that Athena should be aware that she is a robot and not a human. While we recognized a desire in users to engage in conversations where the first person narrative is commonly invoked, and in conversations with sharing of thoughts, feelings, and human experiences, we decided  to limit  Athena to the types of experiences that an Alexa device is capable of. 
%capable of discussing to those that an Alexa device For example, Athena does not discuss dream experiences.

\subsection{System Overview }
Athena is built using the Alexa Skills Kit (ASK)\footnote{\url{https://developer.amazon.com/en-US/alexa/alexa-skills-kit}}, and run as an on-demand application that responds to ASK events containing utterance hypotheses produced by Amazon's automatic speech recognition (ASR) service. Athena's responses are uttered using the text-to-speech (TTS) service provided by the ASK API.  
%Cobot provides several benefits seamless integration with Amazon Web Services (AWS), including AWS Lambda which hosts our application. 

%\begin{figure*}
%\includegraphics[width=0.5\linewidth]{athena-sys-architecture.jpg}
%  \caption{Standard case.}
%  \label{fig:set1}
%\end{figure*}

We built Athena using the Cobot Toolkit Framework~\cite{khatri2018advancing}. 
Cobot provides seamless integration with Amazon Web Services (AWS), and natively utilizes the 
AWS Lambda, DynamoDB, and ECS services. Cobot includes a default socialbot that is an empty incomplete shell of a dialogue system. We used this default socialbot as a launching point for the Athena Dialogue System.  
We refer the reader to \citep{khatri2018advancing} for a detailed description of the Cobot framework and what it includes. 
%We used several AWS services in addition to those included in Cobot by default. Neptune was used to build a knowledge graph, and Amazon Elasticsearch for information retrieval and response retrieval.

\begin{figure}[t]
  \centering
  \includegraphics[width=0.4\textwidth]{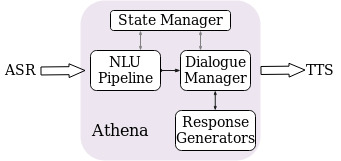}
  \caption{A concise view of Athena's architecture.}
  \label{fig:system-architecture}
\end{figure} 

%\textbf{System Architecture Overview.} 
Figure~\ref{fig:system-architecture} shows a concise view of Athena's architecture.
The inputs to Athena are the ASR hypothesis for a user's turn, as well as conversation identification information. The conversation history and state information are retrieved from a back-end database by a State Manager. Then, the ASR hypothesis is fed into a natural language understanding (NLU) pipeline to produce a collection of NLU features for the user utterance and conversation context. Based on the NLU features and conversation context, a dialogue manager dispatches a call to response generators to populate a response pool. The dialogue manager processes the response pool using a ranking function to select a response to utter next.

%\textbf{Knowledge Base.}
%\textcolor{red}{responsible:  Wen.. Not sure where to put this section}
Large knowledge bases are essential for creating an intelligent and versatile conversational agent \cite{fang2018sounding18,chen2018gunrock}.
Athena uses two knowledge graph resources to aid in Named Entity Resolution (Section \ref{sec:ner}) and knowledge graph-based response generators (Section \ref{sec:kg-based-rg}). 
Athena uses Amazon Information Knowledge Graph (AKG) which covers domains such as movies, music, and books, and is accessed using Alexa Information Query Language within Cobot. 
We use Amazon Neptune to construct an additional knowledge graph from 
English Wikidata\footnote{https://github.com/wikimedia/wikidata-query-rdf}
of 9/2019 \cite{vrandevcic2014wikidata}.
%--- a free, multilingual, and collaborative data collection. 
%We download the Wikidata dump\footnote{https://dumps.wikimedia.org/wikidatawiki/entities/} 
%We use the September 2019 Wikidata dump (English only). %\footnote{https://github.com/wikimedia/wikidata-query-rdf}
 
%responsible: Wen.. Not sure where to put this section We believe that knowledge base is essential to
%create an intelligent and versatile conversational agent. Athena use Amazon Information Knowledge
%Graph (AIKG) which covers domains such as Movie, Music, Politics etc; and it can be accessed
%using Alexa Information Query Language within Cobot. We supplement AIKG with Wikidata [13]
%which is free, multilingual and collaborative data collection. We downloaded the Wikidata dump 2 as
%of September 2019 and prepossessed the dump to only keep English version 3 . We then loaded the
%data into Amazon Neptune 4 and accessed it using SPARQL query language. These two knowledge
%bases are mainly used in Entity Linking and knowledge graph-based response generators that are
%described in detail in Section 4.4 and Section 6.1 respectively.

% this should be deleted? for space
%In the following sections we describe Athena's individual components in %greater detail. 

\section{Natural Language Understanding}
%This section focuses on the NLU modules which we consider most important to Athena's inner workings. 
Athena's NLU configuration follows the general NLU pipeline format provided by Cobot. It runs as a two-stage pipeline where NLU modules within each stage run asynchronously.  We also make heavy use of the NLU services included in Cobot, particularly the topic labeling service and DAIntent taggers. Vader is used for sentiment classification on user utterances~\cite{hutto2014vader}, 
and Spacy\footnote{\url{https://spacy.io/}}  % comment out footnote to save space -davan
is used for part-of-speech tagging. 
%Athena does not employ a coreference resolution model, or perform any ASR correction.

\textbf{Profanity and Red-questions.}
We consider ``red questions'' to be any user utterance which is profane, controversial, or should otherwise be handled in a specific way, such as questions and statements about suicide, financial decisions, or hot button political issues. To detect these utterances, we use an ensemble of the available Cobot topic, intent, and profanity models, in addition to unigram and substring matching. We use a lookup table that contains appropriate responses to various types of red questions. If a red question is detected through a substring match, however, the system will have a more specific and detailed response. 

\textbf{Utterance Segmentation.}
%While ASR improvements have greatly improved Alexa’s ability to correctly understand what the user is saying, %situations where the ASR transcriptions do not seem like a coherent response still arise. However, rather than %being issues with the speech recognition, these errors are largely due to factors stemming from the users %themselves, the environment they are in, and the fact that ASR transcripts do not contain punctuation. In %situations where the user makes a mistake and attempts to correct themselves, repeats a phrase twice, or talks %to someone else in the room, the ASR transcription often picks up all of the words spoken, not just those meant %for Alexa to hear, leading to a noisy ASR hypothesis. 
%We attempt to solve this problem by creating a neural model tasked with detecting sentence boundaries in the ASR %output. To create our training corpus, we hand annotate <N> examples from <CORPUS> with a set of special %punctuation (. , ?) tokens. Treating this as a multi-class, multi-label classification problem, we experimented %with multiple models, including a finetuned BERT based model, BiLSTM-CRF, and Hierarchical CNN(?), with the BERT %model performing best. 
User utterances often contain multiple dialogue acts, but ASR outputs are uncased and do not contain punctuation, thus rendering most pre-existing sentence tokenizers ineffective. There are publicly available utterance segmentation and punctuators available~\cite{tilk2016}, but due to the uniqueness of spoken chit-chat conversations, we  developed our own DA segmentation model. We approach DA segmentation as a sequence classification problem where the goal is to predict the last token of each DA in an utterance, as well as tokens that immediately precede a comma. We use a BERT model initialized with pre-trained weights~\cite{Wolf2019HuggingFacesTS, Devlin_Chang_Lee_Toutanova_2018}. First, the model is trained on the SWDA corpus~\cite{stolcke2000dialogue}. Then, we perform a final fine-tuning on a small corpus of Alexa Prize user utterances that we annotated by hand. Incorporating this model into our system
allowed for increased accuracy in classifying user utterances and intentions by running downstream models on each utterance segment.   

%Examples
%it's going good what about you
%it's going good. what about you?
%i'm sorry i didn't hear you what did you say
%i'm sorry. i didn't hear you. what did you say?
%alexa alexa can you can you tell me about my favorite animal
%alexa. alexa. can you. can you tell me about my favorite animal?

\subsection{Named Entity Linking}
\label{sec:ner}
%\textcolor{red}{responsible:  Wen. input Nikhil Jeff \\ I'm doing an editing pass over NEL section right now.}

Athena's dialogue manager and response generators rely heavily on accurate recognition of named entity mentions in user utterances. Furthermore, entity linking (EL) allows Athena to leverage information stored in large scale knowledge bases, such as knowledge graphs. 
Recognizing named entity mentions, i.e., movie names, musicians, and sports figures, is key to producing coherent and on-topic responses. 
Once entity mentions have been extracted, the entity mentions are linked to their canonical form (or URI) in Wikidata and Amazon Knowledge Graph (AKG). 

In the early stages of the competition we compared SlugNerds~\cite{bowden2018slugnerds} to a number of publicly available off-the-shelf NL and NER tools, namely  DBpedia Spotlight \cite{isem2013daiber},  AIDA \cite{AIDA2011}, and the end-to-end neural entity linker \cite{kolitsas2018end}, after first using truecasing. We found their performance severely lacking. 
Table \ref{table:el_compare}\footnote{We evaluated on recognizing the canonical form of an entity without the entity type since each tools uses a different ontology.} shows the results on a  set of 2000 annotated user utterances. Most of these off-the-shelf tools were trained on editorialized text, such as news-wire, which is very different from the user utterances in spoken dialogue. For example, many NER tools were fitted on cased and punctuated training examples, which causes dramatic performance declines when moving to potentially noisy ASR output that do not contain casing and punctuation. Also, tools such as DBPedia Spotlight were trained on outdated examples from 2016.

Section~\ref{el-ensemble-sec} describes how we developed  an ensemble module that combined DBPedia with a gazeteer produced by retrieving entities from Wikidata and AKG.
We  then trained a new EL system that makes better use of dialogue context, which greatly improved performance (Section~\ref{trained-el-sec}). 

\subsubsection{Entity Linking Ensemble Module}
\label{el-ensemble-sec}
 
%The results of our evaluation  on 2000 annotated user utterances is shown in Table \ref{table:el_compare}.\footnote{We evaluated on recognizing the canonical form of an entity without the entity type since each tools uses a different ontology.} 

%Most of these models were trained on cased text, we, therefore, applied a statistical truecased model to all utterances. \footnote{https://github.com/nreimers/truecaser}
%Table \ref{table:el_compare}
%shows how SlugNerds optimized recall at the expense of precision and that DBpedia spotlight with truecasing performs better than SlugNerds, providing the  best overall performance with an F1 of .51. However, one limitation of  DBPedia Spotlight  is that its entity data is from 2016. In order to increase recall and be able to perform EL for new and recent entities, we created an EL-Ensemble model by combining DBPedia Spotlight with gazetters.

\begin{table}[t]
%\begin{tabular}{p{1.1cm}p{1.2cm}p{1.2cm}p{1.2cm}p{1.3cm}|p{1.2cm}}
 \small
 \begin{tabular}{lllll|l}
\toprule
 & \textbf{Slug.} 
 & \textbf{Spot.} 
 & \textbf{Aida} 
 & \textbf{E2e} 
 & \textbf{Ensbl.}\\ \midrule
\textbf{P}  &   0.33 &   0.46     & 0.61 & 0.64 &\textbf{0.53}\\
\textbf{R}     &   0.66 &   0.57     & 0.17 & 0.07 &\textbf{0.62} \\
\textbf{F1}         &   0.44 &   0.51     & 0.27 & 0.12 & \textbf{0.57}\\
\bottomrule
\end{tabular}
\caption{Performance of entity linking (canonical form only) with existing tools compared to our ensemble module (in last column).}
\label{table:el_compare} 
\end{table}

%\noindent
%\textbf{ElasticNER.}
%\noindent
%\textbf{DBPedia Spotlight.}

In order to increase recall and be able to perform EL for new and recent entities, we created an EL-Ensemble model by combining DBPedia Spotlight with gazetters.
We created gazetteers of 964k entities by querying  AKG for the entity types  Actor, Album, Book, Director, Movie, MusicalAct, Musician, Song, and TvSeries. Additionally, we queried 465k sports-related entities e.g. sports players and sports teams, from Wikidata such as American football, baseball, basketball, soccer, tennis. To enable real-time retrieval, we stored the gazetteers in Amazon Elasticsearch Service,\footnote{https://aws.amazon.com/elasticsearch-service/} indexed by the entity names and types.
%\begin{wrapfigure}{r}{0.4\textwidth}
%\centering
%\vspace{-.05in}
%\includegraphics[width=0.4\textwidth]{figures/entity-type-distribution.png}
%\vspace{-.1in}                         
%\caption{Entity type distribution of DBPedia ontology 
%from April 11\textsuperscript{th} to 20\textsuperscript{th}
%\label{fig:entity_type_dist}}
%\vspace{-.3in}
%\end{wrapfigure}

To guarantee precision: (1) we query the entity candidates by the Elastic Search  score between the entities in the gazetteers and the noun phrases as well as the whole utterance. We rate the candidates by exact match between the whole entities and the text segment in the user utterance; (2) The gazetteers include common phrases, such as “Cool”, and “How are you” as movie names, which increase the false positive rate. We thus created a common phrase list by counting the frequency of entities  in Common Alexa Prize Chats (CAPC) \cite{ram2018conversational}, ignoring entities whose frequency is greater than 60. We manually curated this list to preclude frequent but real entities such as ``Star Wars" or ``Taylor Swift"; (3) We use topic information to restrict search, e.g.  if the topic of the utterance is Music, then we only search  entity types of Album, MusicalAct, Musician, and Song.

%assuming that it's unlikely that a large number of users would mention the same entities coincidentally. Though we find some popular entities such as "Star War" and "Taylor Swift", most of the entities which collected by this method are common phrases. 

% %begin{wraptable}{r}{0.5\textwidth}
% \begin{wrapfigure}{r}{0.5\textwidth}
% %\begin{figure}[h]
% \vspace{-.1in}
% \begin{tabular}{l}
% \toprule
%  \textbf{User:} i really like johnny depp \\ \midrule
%  \textbf{EL output:} \\
%  \textcolor{red}{make this look pretty and forward slash all curly braces. MAW also please fix the size of this.} \\
%  \{ `surface\_form': `johnny depp', \\
% % `canonical\_form': `Johnny Depp', \\
% % `akg\_id': `aie:johnny\_depp', \\
% %      `entity\_type\_schema': {[
% %        `Person'
% %      ]}, \\
% %       `entity\_type\_dbpedia': \\ 
% %       {[`Actor',
% %        `Artist',
% %        `Person',
% %        `Agent'
% %      ]}, \\
% %      `entity\_type\_wikidata': \\
% %      {[`Q5',
% %        `Q483501',
% %        `Q33999',
% %        `Q24229398' \\
% %        `Q215627'
% %      ]}, 
% %       `gender': {[
% %        `male'
% %      ]}, 
% %      `source': `es', \\
% %      `summary': `American actor, film producer, \\ and musician', 
% %      `span': \{
% %          `start': 14,
% %          `end': 24
% %        \}
%       \} \\
      
% \bottomrule
% \end{tabular}
%     \vspace{-0.05in}
%     \caption{An example output of Athena's Entity Linking Module}
%     \label{table:el_output} 
% %\end{wraptable}
% \end{wrapfigure}

\begin{figure}[t]

  %\centering
  \includegraphics[width=0.46\textwidth]{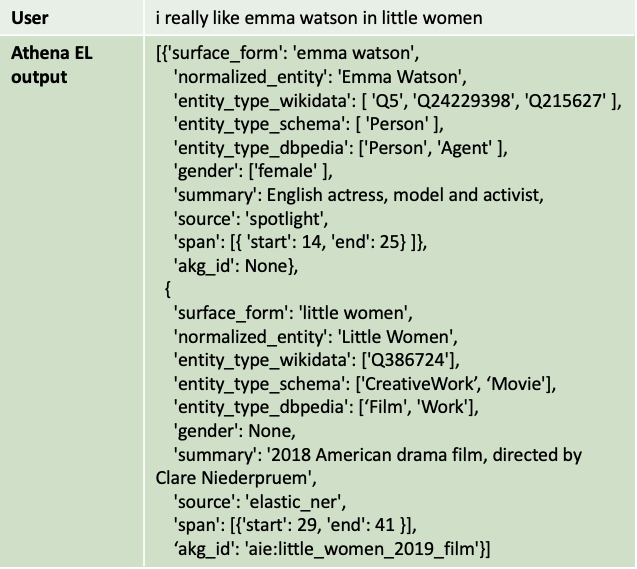}
  \caption{Sample output from EL-ensemble.}
  \label{table:el_output} 

\end{figure}

The resulting EL-ensemble model also uses true-casing, and achieves a large increase in both precision and recall, with an F1 of 0.57 (last column of Table \ref{table:el_compare}). Figure \ref{table:el_output} shows a sample  output of the EL-ensemble module. To supplement this information, we also query our Wikidata database to get gender and a summary for each linked named entity. This facilitates the response generators (Section~\ref{sec:kg-based-rg}) to use the correct  pronoun.  

\subsubsection{Trained Entity Linking Module} 
\label{trained-el-sec}
%In addition, we found that another limitation of the system was that 28.9\% of all the DBPedia entity types returned by Spotlight were missing. Since the DBpedia implementation we were using was trained in 2016, our system performed poorly in recognizing recent entities like latest movies and songs. While the Elastic NER module addressed this issue, it increased false positives considerably.
One of the primary drawbacks of the EL-Ensemble module is a high false positive rate and the fact that we cannot finetune  DBPedia Spotlight. 
%users are frustrated to interact with the system when it is unable to identify named entities %and talk about them, leading to poor user experience and lower ratings. 
To mitigate these problems, we trained an EL system which utilizes contextual information, NLU features, the gazetteers and the two components of the existing EL-Ensemble (Spotlight and ElasticNER). The training consists of three phases.

\noindent
\textbf{Named Entity Recognition.} This is modelled as a sequence labelling task. For a sequence of words of size $m$, $w = (w_1, w_2,...,w_m)$, we identify a sequence of tags $y = (y_1, y_2,...,y_m)$, where $y$ belongs to the set of possible tags. In our experiments, we found the tag set $\{B, I, O\}$ performs best in our set-up. This model allows us to create features that leverage NLU components like topic and dialogue act that are strengths of our system. Furthermore, we use the gazetteers, contextual features, n-grams and the results of our other Named Entity models - DBpedia Spotlight and ElasticNER as input features. We perform training using the perceptron learning algorithm \cite{rosenblatt1957perceptron,collins2002discriminative} and Viterbi algorithm \cite{forney1973viterbi} to find the best sequence labelling assignments.

\noindent
\textbf{Candidate Pool Generation.} Once we have the named entity mentions, we generate a pool of a candidates (up to a maximum size of one thousand) for each entity mention from a knowledge base. We hosted the DBPedia Lookup\footnote{https://github.com/dbpedia/lookup} service and query it to get a ranked list of related DBpedia URIs. In order to get up-to-date information, we re-generated the index for 2017 and 2019 of DBpedia data.  

\noindent
\textbf{Candidate Re-ranking.} The re-ranker learns to rank a pool of candidates and returns the linked named entity with the highest score as a result. We used margin-based loss with various features such as the entity type, the popularity in the knowledge graph, topic, cosine similarity between mention and candidates to train the model.

\noindent
Both models were trained on 20 dialogues from recent user interactions with Athena. Table \ref{table:el_train_performance} shows a significant improvement on both linking to the correct entity and getting the correct entity type on the test data (4 conversations). We also evaluated the existing EL-Ensemble for comparison on the new test set.  We deployed the model towards the end of the semi-final stage and plan to conduct additional evaluation and make iterative improvements with additional training data.

\begin{table}[t]

\centering
\small
%\begin{tabular}{p{3cm}p{3.5cm}p{1.2cm}p{1.2cm}p{1.2cm}}
\begin{tabular}{lllll}
\toprule
 \textbf{EL System} & \textbf{Eval. Cat.} & \textbf{P} & \textbf{R} & \textbf{F1} \\ \midrule
 EL-Ensemble & entity & 0.39	& 0.58	& 0.47 \\
 EL-Ensemble & entity+type & 0.11	& 0.17	& 0.13 \\
 Trained EL & entity & 0.60	& 0.55	&0.55 \\
 Trained EL & entity+type & 0.50	& 0.42	& 0.45 \\
\bottomrule
\end{tabular}
\caption{Evaluation of trained EL.}
\label{table:el_train_performance}

\end{table}

\subsection{Dialogue Act Ontology and Automatic Recognition}
%\textcolor{red}{responsible:  Davan. input Rishi Lyn}

%%%% Dialogue Act Ontology
Dialogue Act (DA) recognition is a key NLU feature that enables different components of the system
%to utilize knowledge of the current speech act and user goal 
to constrain their actions and behavior.  The system currently uses an ensemble of taggers to enable DA recognition. We first evaluated the use of the MIDAS DA scheme \cite{yu2019midas} and tagging model.\footnote{https://github.com/DianDYu/MIDAS\_dialog\_act} We found that there were several limitations with directly using MIDAS in Athena, in particular some DAs in MIDAS, such as \textit{command}, are too coarse-grained to support explicit conversation control, and 
in addition the trained model did not perform as well on dialogues collected with Athena, perhaps because aspects of the model were tuned to the Gunrock system.   

% Motivation for expanding the scheme
We thus developed an expanded DA schema, informed by MIDAS, but with additional distinctions important for dialogue management in Athena. We took 300K utterances from recent CAPC distributions and labelled these user utterances with this schema.
We then developed an  ensemble DA tagger, comprising  the MIDAS BERT model, an SVM DA model, Cobot Intent classifier API, and a  Regex tagger.  A linear SVM classifier was then trained on this expanded scheme using 2,3,4-grams as input features to label each segment of the user utterance. We  also implemented a regex-based tagger that matched phrases in the user utterances, which works well for some types of DAs, but lacks flexibility for cases such as \textit{request-repeat} where there are many ways to phrase a user repetition request.  

\begin{table*}[t]
    \centering
    % \begin{minipage}
    % \end{minipage}
    \begin{small}
    \begin{tabular}{c c}
    \begin{tabular}{l c c c }
        \toprule
        \textbf{Label} & \textbf{P} & \textbf{R} & \textbf{F1}\\
        \midrule
         more-information & 0.73 & 0.57 & 0.64 \\
         change-topic & 0.98 & 0.79 & 0.88\\
         avoid-topic & 1.00 & 1.00 & 1.00 \\
         discuss-topic & 0.98 & 0.99 & 0.98 \\
         signal-non-understanding & 0.99 & 1.00 & 0.99 \\
         personal-question & 0.97 & 0.98 & 0.98\\
         experience-question & 1.00 & 1.00 & 1.00 \\
         request-options & 0.96 & 0.84 & 0.90 \\
         \bottomrule
    \end{tabular}
         &  
    \begin{tabular}{l c c c}
    \toprule
        \textbf{Label} & \textbf{P} & \textbf{R} & \textbf{F1} \\
        \midrule
         advice-question & 0.97 & 0.98 & 0.98 \\
         fact-question & 0.98 & 0.98 & 0.98 \\
         no-answer  & 0.73 & 0.50 & 0.59 \\
         yes-answer & 0.90 & 0.64 & 0.75 \\
         acknowledgement & 0.86 & 0.71 & 0.77\\
         apology & 0.96 & 0.94 & 0.95 \\
         complaint & 0.97 & 0.95 & 0.96 \\
         conversation-closing & 0.94 & 0.90 & 0.92 \\
         \bottomrule
    \end{tabular}
    \end{tabular}
    \caption{Results of SVM Dialogue Act Tagger}
    \label{table:svm_dact_results}
    \vspace{-0.2in}
    \end{small}
\end{table*}
% Image based table commented-out since full table is ready
% \begin{table}[h]
%     \centering
%     \includegraphics[scale=0.25]{figures/svm_dact_results.png}
%     \caption{Results of SVM Dialog Act Tagger}
%     \label{table:svm_dact_results}
% \end{table}

The results for the classification of key DAs are in Table \ref{table:svm_dact_results}. 
Since the DAs form a very disparate group coming from different systems, DAs with similar intent are grouped together. Athena's DA schema distinguishes more types of user questions than previous work, and uses them in the dialogue manager. Also  note that user requests related to changing, avoiding or discussing a topic are generally recognized with extremely high accuracy, and this is thus one of the strengths of Athena.

%
%
% Dialogue Manager
%
\section{Athena's Dialogue Manager}
%lyn commented this. In this section we will describe The components of Athena’s dialogue manager, %but first we will talk about what we think makes a good thing to say next. Also, we will discuss the %structure that we impose on Athena‘s response.

\begin{figure*}
  \centering
  \includegraphics[width=0.8\textwidth]{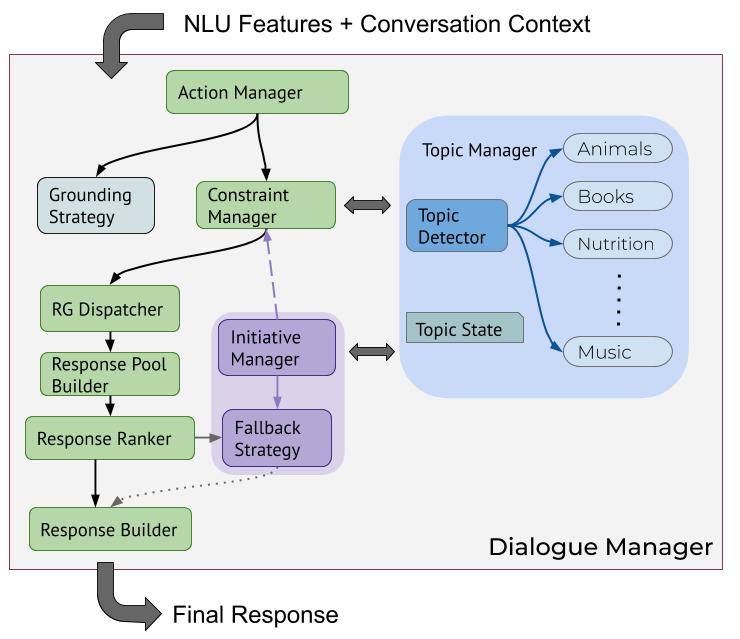}
  \caption{Dialogue manager architecture.}
  \label{fig:dm-architecture}
\end{figure*}

Dialogue management in open conversation is a particularly challenging task  due to the universe of possible valid things that can be said next at any point in  conversation. While goal oriented dialogues have a clear objective which the dialogue manager can optimize when making decisions throughout a conversation, the dialogue manager for open domain dialogues does not have an obvious way to measure the appropriateness of a possible candidate response. We view the lack of a clear and focused criteria in gauging response quality as a primary source of difficulty in developing an open domain dialogue system.

We aimed for a flexible topic agnostic approach to dialogue management that dynamically configures a dialogue based on theories of discourse coherence. Principles of centering theory, topic continuity and discourse relations provide some guidance~\cite{GroszCentering, Prasadetal10Exploiting},
as well as Gricean maxims of quantity, quality, relation, and manner \cite{Grice57}, and observations about mixed-initiative dialogue \cite{WW90}. A good response is one that is interesting and related to previous concepts and entities mentioned by the user or system. The dialogue system should avoid repetition, and mix initiative with a variety of dialogue acts throughout the conversation.  The dialogue manager does conversation planning primarily oriented around statements or questions related to entities in focus, and should use a mixture of opinion and non-opinion content.  The purpose of a question is to collect information that will drive the conversation forward by setting up future statements or questions.  But these principles 
do not provide clear guidance when navigating the low level details of dialogue management decision making.

Athena’s dialogue manager architecture uses a shallow hierarchy based on a top down decomposition of the dialogue management task. The dialogue manager can be decomposed into a number of sub components, each with its own responsibility. The components correspond to phases of dialogue management, and they are oriented as a pipeline: the outputs of one component directly feed as inputs  to the next. The DM sub-modules are given in Figure~\ref{fig:dm-architecture}, and discussed below. 

%DM Sub-components:
%\begin{itemize}
%    \item Topic Manager: these are our topics
%    \item Action Manager
%    \item Grounding Strategy
%    \item Constraint Manager -
%    \item RG Dispatcher - decides which RGs to run given the DM constraints. 
%    \item Response Pool builder - builds a pool of responses. Runs RGs to make a response pool. Filters response pool using profanity + repetition filters.  
%    \item Response Ranker - Backing off through progressively less restrictive response selection criteria. approach to response ranking using heuristics. 
%    \item Fallback Strategy - select new topic, select pre-computed response. 
%    \item response builder - assembles response parts. Some post processing / cleaning.
%    \item Initiative Manager
%\end{itemize}

\begin{table*}[t]
\centering
\begin{small}

\begin{tabular}{p{5.2in}} \toprule
 \texttt{ATHENA:} \textcolor{magenta}{Right.} 
 \textcolor{blue}{I would really like to go hiking in the woods.}
\textcolor{black}{What do you think is the most beautiful thing about nature?} \\ 
\midrule
 \texttt{USER:} when we go camping and the moons out and it's quiet
\\ \midrule
 \texttt{ATHENA:} \textcolor{magenta}{Ah, the wolves? Hmm.}
 \textcolor{blue}{Yes. I think we really need to preserve the beautiful nature around the world as much as we can.}
\textcolor{black}{There are some really beautiful national parks in the U.S.! 
Have you been to the Glacier National Park by any chance? It's in Montana.}
\\ \bottomrule
\end{tabular}
  \caption{Sub-conversation illustrating Athena's response parts construction. The {\sc ground} is magenta, the response {\sc opener} is blue, and the response {\sc body} is in black.}
  \label{table:response-parts-construction}

\end{small}
\end{table*}

\noindent
\textbf{Response Structure.}
Its common for system responses to contain multiple speech acts. Athena’s responses can be decomposed into discrete parts that perform different dialogue purposes and are operated on separately. The subparts are called {\sc ground}, {\sc opener}, and {\sc body}. The example Athena response in Table~\ref{table:response-parts-construction} indicates each  subpart. 
The {\sc ground} always comes first:  it is limited to a strictly backward looking grounding dialogue act \cite{stolcke2000dialogue}. The {\sc opener} is dynamic and can be used for several purposes. It can be forward or backward looking.  The opener may be used to smooth the transition from one topic to another. It is commonly used to set up what will be said next by noting how it’s relevant: either directly to things previously said in conversation, or to the current topic. The response {\sc body}  contains new information that drives the conversation forward, and may contain multiple dialogue acts.  For example the {\sc body} may include a {\sc hand-off} as described in Section~\ref{sec:flow-rg}. The dialogue manager tracks the response parts separately.

\begin{figure*}
  \centering
 \includegraphics[width=0.8\textwidth]{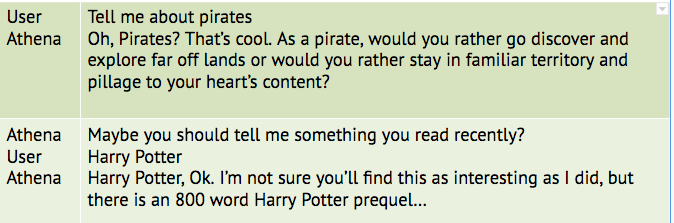}
  \caption{Examples of grounding phrases.
 \label{grounding-fig}}

\end{figure*}

\noindent
\textbf{Grounding Strategy.}
Given the response structure, every system utterance potentially includes a {\sc ground}
generated via a grounding module.  Examples are shown in Figure~\ref{grounding-fig}. This serves two purposes; (1) grounding utterances are sent as progressive responses reducing system response delay; and (2) grounding makes the utterances more natural by following conventions of human-human dialogue, and increase the user's confidence that the system has understood them correctly \cite{fang2018sounding18,Traum94}. As shown in Figure~\ref{grounding-fig}, grounding can provide a confirmation of Athena's understanding of a topic or named-entity, as well as provide evaluative feedback to the user, e.g. {\it That's cool}.
In order to send the grounding phrases as progressive responses, our grounding strategy is completely backwards looking, conditioned on dialogue act, and independent of the topic. For any given dialogue act, the baseline response checks for named entities, and uses templates to return a short, general repetition of this entity. However, for some specific dialogue acts, such as “command”, “opinion”, and “comment”, the system uses intent classification and sentiment models, in addition to regex matching, to detect if a more finegrained grounding can be returned. These responses include more specific templates, in the case that a named entity was recognized, as well as templates that do not require NE’s, such as pure backchannels and acknowledgements of commands. 
We conducted an A/B test of grounding when it was first deployed with  results indicating that grounding improved user ratings.

\noindent
\textbf{RG-DM Interface.} 
The interface between the dialogue manager and response generators is a contract-based approach. The dialogue manager passes a set of response conditions to the response generators. The conditions act as a contract that the response generators must meet in order for their response to be considered for the current turn. This approach gives our dialogue system the flexibility to take advantage of a wide variety of response generator types, such as indexing and retrieval RGs, generative response models~\cite{harrison2019maximizing, oraby2019curate}, 
and call flows that resemble finite state automata. 
The DM-RG interface allows for backwards compatibility, e.g.  Athena supports a legacy system by treating it as an RG. In addition, completely swapping out all of Athena's RGs would require little modification to Athena's dialogue manager, consisting of updates to the response ranking function. 

Response generators can return multiple response candidates at each turn. In fact, this is encouraged in case some responses get removed during a quality filtering stage. If an RG does not have anything of substance to say at a given point in conversation, then it returns a null response (rather than saying something along the lines of ``I don't know what that is''). Furthermore, RGs are required to return the response parts as labeled separate segments.

\begin{table*}[t]

\centering

    \begin{tabular}{p{1.1in}|p{3.8in} } \toprule
        \textbf{System action} & \textbf{Description} \\ \toprule 
        \texttt{perform\_repeat} & Repeat the previous turns utterance. \\ \midrule
        \texttt{conv\_closing} & End the conversation. \\ \midrule
        \texttt{advise\_usage} & Provide usage instructions. \\ \midrule
        \texttt{greet} & Start a conversation. \\ \midrule
        \texttt{repeat\_request} & Request that the user repeats themselves. \\ \midrule
        \texttt{wait\_prompting} & Wait for the user to think/finish what they were saying. \\ \midrule
        \texttt{red\_response} & Respond to a "red question". \\ \midrule
        \texttt{topic\_change} & Initiate a change of topic. \\ \midrule
        \texttt{list\_options} & Provide a verbal menu of options. \\ \midrule
        \texttt{converse} & Everything else. \\ 
         \bottomrule
    \end{tabular}
  \caption{Athena's system actions. }
  \label{table:dm-actions}

\end{table*}

\noindent
\textbf{Action Manager.}
Situations commonly arise where the dialogue system needs to facilitate the conversation by performing some functional speech act. These situations are topic independent and may occur at any point in dialogue. The job of the Action Manager is to recognize these situations and decide on an appropriate course of action for the system to take. We use the set of nine system action types in Table~\ref{table:dm-actions}. The \texttt{converse} action denotes the case where a functional speech act should not be performed. The Action manager uses a heuristic rule method based on NLU features, such as dialogue act, keyword matching, and regular expressions, to decide the next system action.

\noindent
\textbf{Constraint Manager.}
Once the system action has been decided, the Constraint Manager is responsible for generating a set of constraints that describe the next system utterance. The constraints specify a topic of discussion, as well as potential entities to mention. Also, the constraints contain a flag that signals if the dialogue manager is initiating discussion on a new topic this turn.  For example, the constraints: \texttt{\{ Topic = harry\_potter, Entity Mention = J.K\_Rowling,  Dialogue\_act = opinion\_question \}}, 
describe the following system utterance:
“\textit{What do you think about J.K. Rowling? Do you know much about her?}”

Currently, the constraint generation is based on heuristic rules. The constraint manager is designed using a flexible topic-agnostic approach  based on general principles of entity/topical coherence. In short, the general strategy of the constraint manager is to talk about the same entities or concepts as the user. The constraint manager can specify soft or hard constraints. The dialogue act is a soft constraint, and we attempt to use a variety of dialogue act types to increase engagement. Primarily, the DM alternates between opinion statements, non-opinion statement, and opinion questions.    

In the process of generating constraints for the next turn, the Constraint Manager asks some internal questions: What is the current topic? Is it a continuation from the last turn's topic or has the user changed it to something new? Are we still on the same topic, but have shifted to a new sub-topic, or focused on a new entity? These questions are answered by another module: the Topic Manager sub-module, described later in this section. 

While generating constraints, the DM has the ability to initiate a topic change action. Two of the most common reasons for changing the topic during this phase are if the DM loses track of the current topic, or does not know appropriate constraints to produce for current conversational context.
Conversation constraint generation is one area that we are very interested in from a research perspective, and we are currently  constructing a machine learning model based approach to constraint generation.

\noindent
\textbf{Topic Manager.}
%\textcolor{red}{Describe Topic manager by writing out information from the Amazon site visit presentation slides on topic manager.}
The Topic Manager refers to a collection of modules and sub-modules that perform several tasks related to topic representation, topic detection, and topic state representation.  Athena's dialogue manager uses a topic state to track topical information related to the current conversation so far. In addition to the current topic, the topic state includes a distribution over turns spent on each topic so far, and a topic history containing the sequence of topics visited. It contains lists of entities mentioned by both the user and the system. That being said, most of Athena's RGs do not supply responses with annotated entity mentions.   

Individual topics are represented using Python classes. Each class stores information related to that topic such as its name, referential expressions associated with the topic, and subtopics that fall within its general domain, e.g., basketball is a subtopic of sports. The arrangement of topics is designed to reflect Athena's conversational needs rather than real world topical relationships. For instance, one might consider artificial intelligence a subtopic of science and technology rather than a separate topic, which is how they are represented within Athena. Table~\ref{table:athena-topics} lists all the topics that Athena is capable of discussing. Some topics, such as \textit{politics} and \textit{controversial}, have topic classes implemented within Athena so that she can respond appropriately when a user is trying to talk about them. 

\begin{table}
    \centering

    \begin{tabular}{ll}
        \toprule
        % \multicolumn{2}{c}{Topics} \\
        % \midrule
        \textit{sports} & \textit{movies} \\
        \textit{books} & \textit{nature} \\
        \textit{news} & \textit{animals} \\
        \textit{astronomy} & \textit{comic books} \\
        \textit{dinosaurs} & \textit{harry potter} \\
        \textit{nutrition} & \textit{pirates} \\
        \textit{video games} & \textit{board games} \\
        \textit{hobbies} & \textit{science and} \\
         & \textit{technology} \\
        \textit{introduction} & \textit{persona} \\
        \textit{controversial} & \textit{politics} \\
         %&  \\
         \bottomrule
    \end{tabular}
  \caption{Athena's topics.}
  \label{table:athena-topics}

\end{table}
The topic detector module also falls under the purview of the Topic Manager. 
The topic detector searches the user utterance for topic name mentions and key words associated with each topic. Then, a combination of information pieces, primarily dialogue act and topic key-words, are used to recognize and detect situations where the user is trying to invoke some new topic. We categorize user topic invocations into two types. In the first type, the user explicitly invokes the topic by name, e.g., "let's talk about sports". In the second type, the user might only refer to some entity related to Athena's internal topic, e.g., "what do you think about Kobe Bryant?" falls within the sports topic class. Athena's contract based dialogue management approach requires that there is a discrete topic class assigned to each conversation turn. Therefore, correctly identifying entities and linking them to their associated topic is paramount.

\noindent
\textbf{Initiative Manager.}
In some situations the DM may decide to change the topic of discussion. The Initiative Manager comes into play after the decision has been made to leave the current topic, and it is responsible for deciding what new topic to initiate conversation on, or whether to give the user the initiative.

\noindent
\textbf{RG Dispatcher.}
The RG dispatcher decides which RGs to run given the action and constraints. In most cases Athena has two or three RGs that can contribute to conversations within a given topic.  All RGs capable of meeting the constraints are selected. Also,  some RGs run every turn, such as  RGs for responding to “red” questions, RGs that contain Athena’s persona responses, and question-answering RGs such as Evi. 

The decision making in the response dispatcher is an automatic process mainly achieved through look up tables. We construct a registry by requiring each RG to register (at time of instantiation) the action types and topics that they are capable of responding to. Then, during execution, the registry determines which RGs to run based on the outputs of the Action Manager and Constraint Manager.

\noindent
\textbf{Response Pool Builder.}
The response pool builder is the DM module that interfaces with the response generators. This module takes the list of RGs output by the RG Dispatcher and calls on them to populate a preliminary response pool. This preliminary response pool is refined with filters for profanity and repetition and then gets sent back to the DM. Before passing the response candidates to Cobot's offensive speech classifier, we mask words and phrases that we observed to cause responses to be flagged falsely as offensive, such as ``king'', ``president'', or ``saturated''.

\noindent
\textbf{Response Ranker.}
The goal of the response ranker is to find the best response that also matches the constraints output by the Constraint Manager. 
Athena has two response rankers, one of which is largely based on heuristic rules that back off through progressively less restrictive response selection criteria, and the other which is trained from Athena conversations.

\noindent
\textbf{Fallback Strategy.}
Several circumstances can lead to the response pool being empty. For instance, a system crash,  the RGs could not meet the dialogue manager constraints, or all the responses were removed from the response pool during the filtering stage. The purpose of the Fallback Strategy module is to take over and provide a graceful way of moving the conversation forward.  Our most frequent fallback approach is to initiate conversation on a new previously unvisited topic, but after several system initiatives, Athena will use a fallback strategy that gives the user the initiative by prompting for an unvisited topic.  The Fallback Strategy selects a template utterance from a collection of utterances crafted specially for this circumstance.  

\noindent
\textbf{Response Builder.} 
The Response Builder module assembles the response parts and performs a cleaning post-processing step, before sending the response to the TTS engine.

%
% Response Generators
%
\section{Response Generators}

To enable scalable dynamic dialogue interactions, our main aim was to develop multiple response generators for each topic, that provide different types of dialogue turns and generic methods for managing topical transitions.

\textbf{Centering-Based RGs.}
Athena's centering-based RGs are retrieval based generators motivated by theories of discourse centering and topic continuity\cite{GroszCentering}. The primary discourse relation performed by this type of RG is Elaboration, and responses focus on entities and topic specific concepts. This RG relies on a response bank containing utterances that are annotated for entity mentions and other contextually relevant information. This type of RG is paired with an NLU component that automatically recognizes entities and concepts mentioned by the user. The response retrieval is carried out using a heuristic scorer that looks for the response candidate most similar to the user's most recent concept and/or entity mentions. 
Stylistically, centering RG responses typically contain some factual information, such as a fun-fact, that is either preceded or followed by some opinions or subjective evaluative language.  

%The discourse relation is Elaboration, entity-based coherence.

\textbf{Elasticsearch Retrieval.}
%\textcolor{red}{Jiaqi. input from Jurik} 
We utilize Elasticsearch to retrieve responses curated from Reddit and labelled with named entities \cite{gopalakrishnan2019topical}. We create Elasticsearch Indices for 10,969 entities and responses pairs. NER outputs are used to query the response index. To make it sound more natural, the response begins with "I wonder if you know that ...". 
%For example, if the user mentioned the "Google" entity, we will return candidates like "I wonder if you've heard this. Google Maps calculates traffic by tracking how fast Android devices are moving on the road."

\textbf{Back Story.}
The backstory response generator responds to questions about the bot's preferences. The idea is to create a consistent persona that is curated by the team. We use regex string matching to identify common questions users ask the system. 
%using the CAPC~\cite{ram2018conversational} data. 
These questions can be termed as favorite questions, i.e.,
\textit{what is your favorite TV series?} or \textit{what type of food do you like the most?} 
%We calculate the BLEU score of the utterance with each candidate and return the response is the score is higher than the 0.7 threshold.

%\subsection{News}
%The News RG allows the bot to converse about current events. It sources data from Washington Post articles and generates a three-turn conversation about a relevant article. To ensure that the conversations stay current, we store the hundred most recent articles in every deployment. Furthermore, we filter the articles based on a curated list of topics. The RG selects the articles using the entities identified by the NER model from the NLU pipeline and matches them against the headline and the themes of the article. If the user responds positively or expresses the intent to discuss more, the RG continues to the follow-on turns. In the initial turn, the News RG introduces a relevant article using the headline. In the following turn, the entities linked with the article are described to inform the user of the context of the article. Finally, the article is summarized using an off-the-shelf summarization model \footnote{https://pypi.org/project/pysummarization/}. At any turn, if the user does not express interest in the article, a new article is selected to continue the discussion. 

\textbf{News.}
The News RG allows the bot to converse about current events. It sources data from Washington Post articles and generates a three-turn conversation about a relevant article. To ensure that the conversations stay current, we store the hundred most recent articles in every deployment, after applying topical filters and summarizing them. \footnote{https://pypi.org/project/pysummarization/}
%Articles are selected by matching the headline and the themes of the %article to NER module outputs. The conversation continues for up to three %turns as long as the user responds the user responds favorably. 
%First, an article is introduced using its headline. Next, contextual %information is given by describing the entities linked with the article 

%At any turn, if the user does not express interest in the article, a new article is selected to continue the discussion.

\textbf{Neural Response Generators.}
\label{rg-sec}
We integrated the Topical Chat Neural Response Generator (TC-NRG) into Athena and tested it with knowledge retrieved in context from either Wikipedia or the Amazon Knowledge Graph. Section~\ref{tc-nrg-sec} reports the results of an evaluation of the generated responses.

% In addition to TC-NRG, we have recently started the development of a hybrid RG based on a neural model, specifically for the video game topic. The model is trained on the ViGGO dataset~\cite{juraska2019viggo}, and it differs from TC-NRG in that it generates responses from structured meaning representations. The model's input is thus an explicit indication of what the content and intent of the response should be, as opposed to determining it on its own from the dialogue context. We therefore wrap it in our Flow-RG framework for call-flow-style RGs (see Section~\ref{sec:flow-rg}), which will take care of generating these response specifications based on the current context.

\textbf{Text-to-Speech Synthesis.}
Since the Alexa platform provides the text-to-speech synthesis capability, Athena only needs to produce responses in textual form. In order to make the responses sound more natural, however, we make use of Speech Synthesis Markup Language (SSML). This allows us, among other things, to (1) slightly reduce the pronunciation rate of longer responses containing facts, (2) incorporate human-like interjections into the responses, especially their beginnings, and (3) have Athena react with more emotion when the occasion calls for it.

In our experience, the SSML often makes interjections inappropriately expressive, so we limited its use to only a few particular ones. Adding the exciting emotion using SSML, on the other hand, helps many responses sound more engaging. We observed, however, that whether they sound natural or not, depends on the exact wording of the response. We therefore abstain from using automatic methods to inject SSML that modifies emotion, and instead we opted for manual addition to individual response templates after first verifying the pronunciation in the Alexa Developer Console. To make the manual annotation process faster and less error-prone, we wrote a range of parametrizable methods that generate and inject SSML code into text that can be utilized in any RG.

\subsection{Knowledge Graph-based Generators}
\label{sec:kg-based-rg}
%\textcolor{red}{responsible: Lena. input from Wen Lena Kevin}

Generating responses directly from knowledge-graphs has two possible advantages; (1) it provides a highly scalable source of content, and (2) it provides a natural structure for shifting to a related topic or entity by following relational links in the graph.\footnote{We also show in Section~\ref{tc-nrg-sec} that information from the knowledge graph seems to be effectively used by the topical chats neural response generator.} We hypothesized that the ability to shift to a related entity would support deeper and longer topical conversations that would be coherent
by virtue of the links between related entities. 

We use knowledge graph-based response generators for the music, movies and sports topics. Movies and music 
utilize the Alexa Information Knowledge Graph (AKG) and 
sports uses WikiData. All three response generators use the NER to 
get potential entities from the user's utterance, which we then look-up in the associated knowledge graph. We disambiguate multiple candidate entities using the expected entity type. 

 We  generate three possible types of responses using the links and relations in the graph. 
There are {\sc on topic} responses, 
which try to keep providing information about the entity in focus. 
%MAW. Not all relations are available for all entities and therefore our  generators need to be flexible about responding with a particular  relation. 
There are also {\sc shift topic} responses. These use the links in the KG to
find a new entity related to the current focused entity.
%The response  explains how the new entity is found and asks the user if this new entity is acceptable.
The third type  is a {\sc favorite entity} 
response, which allows Athena to introduce new (popular) entities in order to keep the conversation going on the topic when the user hasn't taken the initiative to do so. 
%MAW Sometimes there are no new relations and the entity cannot  switch, or if a user does not want to continue to the switched to entity, the system can fallback to a list of favorite entities. 
Table \ref{table:convo_kg} provides an example 
conversation with all three types of responses.

% \begin{wraptable}[t!bh]
% \begin{footnotesize}
% \begin{tabular}
% {@{} p{0.4in}|p{2.5in} @{}} \toprule
% ATHENA & Do you have a favorite musician?
%   \\ \hline
% \end{tabular}
%  \caption{Meaning representation and output training pair \label{table:MR}}
% \end{footnotesize}
% \end{wraptable}

\begin{table}

    \small
    \begin{tabular}{@{} p{0.5in}|p{2.4in} @{}} \toprule
        ATHENA & Do you have a favorite musician? \\
        USER & yeah, taylor swift. \\
        ATHENA & Ok, Taylor Swift. Wow! Taylor Swift is very prolific!
        She has 114 songs, that's a lot! \\
        USER & yeah, that is a lot \\
        ATHENA & I like Taylor Swift's song, 22, it came out in 2013. 
        Do you like that song? \\
        USER & yeah, it's a good one. \\
        ATHENA & Right? This is interesting, Taylor Swift sings the 
        song Bad Blood with Kendrick Lamar, want to hear more about
        Kendrick Lamar? \\
        USER & not really \\
        ATHENA & One of my favorite musicians is P!nk. Isn't this
        interesting, P!nk won a Grammy Award for Best Pop Collaboration
        with Vocals. Do you want to hear more? \\
        \hline
    \end{tabular}
    %\end{scriptsize}
    \caption{Example conversation using knowledge graph. \label{table:convo_kg}}
\end{table}

While single relation utterances are the easiest to generate, 
they are also quite simplistic and boring, e.g., "\{entity\} has \{num\}
tracks". We can generate more interesting KG-responses using three methods. First, we combine multiple 
relations together to create more complex sentences. Second, 
we interpret the data provided by the relation, allowing the 
system to give an opinion. Finally, we add questions 
to the end of the response. The techniques applied to the response vary based on the relations used.

Some relations can be used jointly to create a more complex response.
We can combine two related relations, such as "\{entity\} is 
married to \{spouse\} ({\sc isMarriedTo}) and has no children." ( {\sc isAChildOf}). 
We can also chain relations. "\{album\} has \{entity\}'s song, \{song\_name\} on it," retrieves one of entity's albums using {\sc isAStudioAlbumBy} and then retrieves a song of the album using {\sc isAnAlbumTrackOn}. 

We can also create more complex utterances by interpreting the value 
associated with the relation. This works best with numerical data, such as IMDb scores. For example, if an actor's average IMDB score is greater than 6.6 we return the response "I guess in general people must really like \{entity\}'s movies." We are able to generate thresholds, where if the value exceeds, or is below, a certain value, we make a logical assumption about the relation and 
add that to our response.

Appending questions to the end of responses allows the user 
to have something to add to the conversation. We can ask questions 
such as, "Do you like this movie?", or "Have you heard of this album?".
When questions such as these are added to the end of an utterance it 
is important to tailor our subsequent responses accordingly. 
% To do this we keep track of the previously asked question to allow us to
% tailor our following utterance to acknowledge their answer. 

% Here is an example conversation:
% \begin{itemize}
%     \item ATHENA: Do you have a favorite musician?
%     \item USER: yeah, taylor swift.
%     \item ATHENA: Ok, Taylor Swift. Wow! Taylor Swift is very prolific! She has 
%     114 songs, that's a lot!
%     \item USER: yeah, that is a lot
%     \item ATHENA: I like Taylor Swift's song, 22, it came out in 2013. Do you like that song?
%     \item USER: yeah, it's a good one.
%     \item ATHENA: Right? This is interesting, Taylor Swift sings the song Bad Blood 
%     with Kendrick Lamar, want to hear more about Kendrick Lamar?
%     \item USER: not really
%     \item ATHENA: One of my favorite musicians is P!nk. Isn't this interesting, P!nk won a 
%     Grammy Award for Best Pop Collaboration with Vocals. Do you want to hear more?
% \end{itemize}

Working with the knowledge graphs can generate interesting responses 
and conversations, but there are several challenges which must be addressed to use them effectively - finding the correct entity, finding interesting relations that are not too sparse, and confirming the data is up-to-date. It is important to make sure that the system does not 
retrieve the entity for Chris Evans the politician when the user wants 
to talk about Chris Evans the actor, but sometimes this can be 
difficult to achieve, especially within topics with many overlapping 
names, such as song titles. There are many interesting facts that can 
be said about an entity, but spending time crafting versatile templates for each relation requires a lot of manual effort. Future work should explore using a trained NLG in combination with large pre-trained language models to generate from the knowledge graph \cite{moon2019opendialkg,hedayatnia2020policy,reedetal20}. Querying the knowledge graphs for each relation can also be time consuming, and increase response delay. %Also, the relation must be queried to determine if it is populated,
% having many unpopulated relations means that it may take the system 
% too long to find a relation that can be used. 
Also, using out-of-date 
or incorrect information makes the dialogue system seem unreliable, and must be avoided, e.g. if we are unsure that the 
{\sc won} relation has every single award an entity has won it can be 
better to just present a single award rather than responding with a 
number of awards won.

\subsection{Flow-Based Generators}

Although we acknowledge the limitation of call-flows, especially in their scalability and flexibility, they are by far the quickest and safest approach to giving a socialbot the ability to have a meaningful conversation on a particular topic. The Alexa Prize socialbots are highly customer-oriented products, and it is clear from user feedback that customers would like to be able to talk about many different topics, and at the same time have a mixed-initiative conversation in which the socialbot both asks relevant questions and is able to answer questions on these topics.

Considering the difficulty of the task of training a data-driven language generator that would be coherent and relevant in the context of a given conversation, we  first focused our efforts on developing a handful of call-flows that, combined, would support different topics and would enable the user to have a relatively substantial conversation with Athena from early on in the competition. The result of this effort was (1) connecting a reduced version of our legacy system to Athena, as well as (2) building a new framework for simple development of call-flow RGs.

\textbf{Legacy System.}
In order to leverage our previous experience in the Alexa Prize competition, we enabled a limited number of call-flows in our legacy system, SlugBot \cite{bowden2019slugbot}. SlugBot is hosted on an EC2 instance and accessed by Athena through a Flask application. We leave the technical description of this system to our 2019 technical report \cite{bowden2019slugbot}.

SlugBot contributes content for the following topics: animals, board games, books, comic books, dinosaurs, hobbies, movies, music, pirates, and video games. We selected topics which were most robustly supported, and further iterated their design to match the standards enforced throughout Athena's native RGs. The majority of the content retrieved by SlugBot is organized into several sequences of general topic oriented chit-chat. We additionally enabled a limited number of topic annotated trivia, and personal questions revolving around would you rather questions and hypothetical questions. This content has been shown to successfully extend topical depth \cite{bowden2019cui}. We also use SlugBot's call-flows for other topics, such as nature and astronomy, to guide the recreation of similar topics in Athena's new Flow-RG framework, described below.

\subsubsection{Flow-RG}
\label{sec:flow-rg}

Flow-RG is a framework that we developed with the objective of creating
dialogue-act (DA) driven call-flow RGs that are modular and flexible. The restriction to DAs is intended to enable general, flexible call-flows. The definition of an RG in this framework consists of three main components.
First, a flow graph consisting of dictionary-like nodes, and DA-based edges between nodes. Edges, each associated with one or more DA labels, determine which node of the flow to move on to, given the DA of the user utterance. Edges can lead to previous nodes, creating loops, which can be used to allow a flow to talk about multiple entities, or to ask the user for specifications or to repeat what they said. Nodes specify the response for the corresponding conversation turn. A response can be composed of multiple \emph{segments}, each chosen or generated independently. The second component is response segment templates and their alternatives (paraphrases). 
The third component is callback functions that generate responses (or their segments) that are more context-dependent.

%\begin{itemize}
%    \item A flow graph consisting of dictionary-like nodes, and DA-based edges between nodes.
%    \begin{itemize}
%        \item Edges, each associated with one or more DA labels, determine which node of the flow to move on to, given the DA of the user utterance. Edges can lead to previous nodes, creating thus loops, which can be used to allow a flow to talk about multiple entities, or to ask the user for specifications or to repeat what they said.
%        \item Nodes specify the response for the corresponding conversation turn. A response can be composed of multiple \emph{segments}, each chosen or generated independently.
%    \end{itemize}
%    \item Response segment templates and their alternatives (paraphrases).
%    \item Callback functions that generate responses (or their segments) that are more context-dependent.
%\end{itemize}

A flow graph can be broken down into smaller \emph{miniflows} that are independent and can possibly be executed in an arbitrary order. Each RG then typically handles a single topic, with multiple miniflows being responsible for different subtopics or for more elaborate transitions between subtopics. However, there is nothing preventing an RG from using the miniflows for individual topics, such as we do in our Introduction RG, where each miniflow handles a 2-3-turn exchange on a casual topic. Below, we provide more detail on the strengths of the Flow-RG framework, and how they allow it to be used seamlessly alongside other types of RGs in the system, complementing each other.

\textbf{Flow Execution and Modularity.}
% \textcolor{red}{TODO(Jurik): Create a diagram of Flow-RG.}
In general, the flow execution begins in an initial node that we refer to as the \emph{root}, when the RG is invoked by the DM upon a topic change.\footnote{Note that although the execution of the flow generally proceeds ``downward'' from the root node, the flow graph is not necessarily a tree, as it can contain loops.} A flow graph can define multiple root nodes for different conditions under which the flow can be initiated, the two factors considered being the initiative (\emph{system} vs. \emph{user}) and whether the flow was previously visited.

In each subsequent turn, an edge is followed---based on the user utterance DA labels---to the next node. If multiple edges from one node have the same label, one is randomly picked, which is one of the several sources of randomness ensuring that each subdialogue generated by a flow is different from the previous. Indefinite loops are prevented by imposing a restriction on how many times any one node can be visited, along with a definition of an alternative ``exit'' response in the corresponding nodes for a smooth transition out of the loop.

When an RG has multiple miniflows defined for different subtopics, they can be configured to be executed sequentially or in a random order. The switch to another not yet visited miniflow happens automatically when a \emph{leaf node} is reached in a miniflow's graph. Alternatively, leaf nodes can explicitly indicate which miniflow to switch to, or the execution can be limited to just one random miniflow from the entire pool.

The modularity of the flow definition, along with the configurability of its execution, allows for easy creation of various types of call-flows that feel different in each conversation. Additionally, with the multiple-miniflow design it can be automatically enforced that a previously visited miniflow is avoided when the RG's topic is revisited in the same conversation, unless the initiative comes explicitly from the user. Any arbitrary information from the flow state can be automatically persisted in the database, and later restored and consulted when the flow is revisited.

\begin{figure*}
    \centering
    \includegraphics[width=0.8\textwidth]{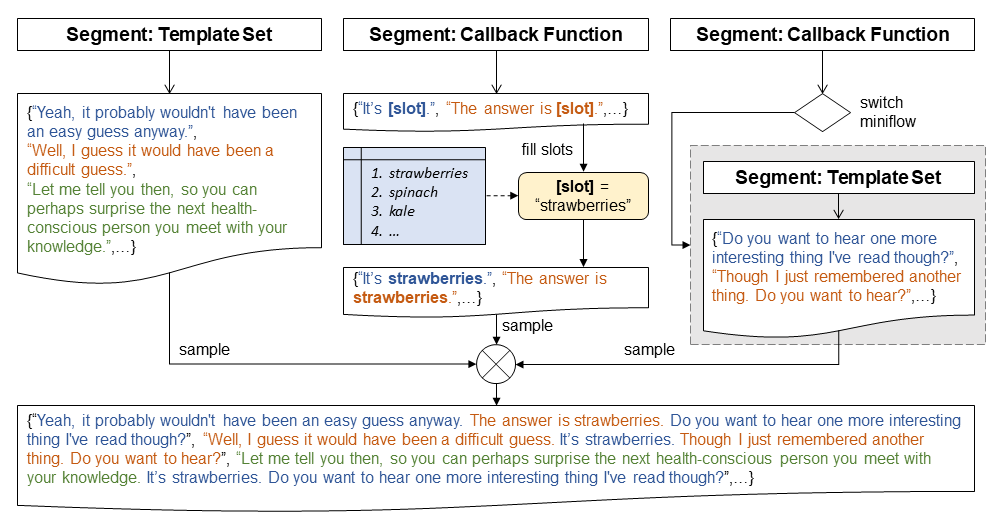}
    \caption{Illustration of response composition in Flow-RG.}
    \label{fig:flow-rg-response-composition}
\end{figure*}

\textbf{Response Composition.}
The response in each turn is assembled from segments specified in the corresponding node. Each segment is defined either (1) in the form of a set of templates, or (2) as a callback function that returns a set of templates. The former offers a simple way of specifying segments in their full form, while allowing alternative templates for increased diversity of the responses. On the other hand, a callback function is more robust in that it can use the previous context and more of the NLU information about the user utterance. It can thus be used to fill in any slots present in the templates, or even to generate or retrieve a segment of a response, based on the user utterance. The two types of segment definitions can be freely combined in the same node.

Figure~\ref{fig:flow-rg-response-composition} shows the process of an example response being assembled from three segments, two of which are different types of a simple callback function: one fills a template slot with a value from the associated knowledge source, while the other initiates a new miniflow and composes the response text recursively, which ultimately corresponds to the last  segment in the example.

When composing a response, each segment's final set of texts is sampled, and all of them are concatenated. This is repeated until up to five different response candidates are composed. These are eventually all returned to the DM, which picks one of them that is not too similar to any of Athena's previous responses.

\textbf{Interweaving with Other RGs.}
In line with the DM's way of response building, the final response in Flow-RG is split into three parts: an opener, a body, and a hand-off.\footnote{This applies to each individual response candidate in the set returned to the DM.} This response structure is particularly useful for creating seamless transitions whether it is between miniflows, or between two RGs. To this end, Flow-RG sets the response from an ending miniflow as the opener (typically, some form of acknowledgement of the user's response, or a short answer), and the body and hand-off parts are reserved for the intro response provided by a new miniflow.
The same mechanism is used for certain transitions from Flow-RG to a different RG, mainly: 
1) when the flow's content is exhausted, and transitions thus to a fallback response chosen by the DM that initiates a new topic, and 2) when a leaf node of the miniflow is reached, and the DM decides to switch to a different RG on the same topic.
%\begin{enumerate}
%    \setlength{\itemsep}{0pt}
%    \item When the flow's content is exhausted, and transitions thus to a fallback response chosen by the DM that initiates a new topic.
%    \item When a leaf node of the miniflow is reached, and the DM decides to switch to a different RG on the same topic.
%\end{enumerate}
The latter is utilized in the DM's interweaving strategy, wherein a flow-based RG takes turns with an entity-centric or fun-fact-based RG in handling a subdialogue on the same topic.

\begin{table}
    \begin{tabular}{p{0.2in} p{2.6in}}
        \textbf{RG} & \textbf{Utterance} \\ \toprule
DRG & Well, I've really enjoyed talking to you about sports. But guess what? I have been working on learning more about Harry Potter. And actually, I'm not sure if my favorite character is Professor McGonagall or Hagrid. What about you? Do you have a favorite Harry Potter character? \\
\rowcolor[HTML]{EFEFEF}
User & yeah i'd say malfoy \\
CRG & Malfoy, ok. I think it's funny that the actor who played Draco Malfoy, Tom Felton, really likes food. While filming the Prisoner of Azkaban movie, the pockets of Tom Felton's Hogwarts robes were sewn shut to stop him sneaking food on set. \\
\rowcolor[HTML]{EFEFEF}
User & that's funny \\
FRG & Definitely. What are your thoughts on this? Would you rather be in Gryffindor, or Slytherin? \\
\rowcolor[HTML]{EFEFEF}
User & slytherin \\
{FRG \newline + \newline CRG} & Slytherin is a good choice! In my opinion, Gryffindors are brave and Slytherins are cunning, I think I'm both so I'll let the hat choose for me. What did you think about the actor Tom Felton, who played Draco Malfoy in the movies? I heard that he originally auditioned for the roles of Harry and also Ron.
\\ \bottomrule
\end{tabular}
    \caption{An example Athena conversation illustrating interweaving RG behavior.}
    \label{table:interleaving-conversation2}

\end{table}

Flow-RG makes it possible for a flow to resume after a few turns handled by a different RG on the same topic. In the above scenario, it simply begins a new miniflow, if there is at least one remaining that has not been visited yet. However, resumption is also possible in the middle of a miniflow, which allows a different RG to chime in for up to two turns (such as Evi answering an on-topic factual question that the flow has no answer prepared for), and then have the miniflow pick up where it left off. Admittedly, the latter might not always result in a smooth continuation, but we are planning to further optimize this strategy.

\textbf{Introduction RG.}
Athena's introductory turns in a conversation are primarily handled by the Introduction RG, which is designed to greet the user and subsequently talk about a few casual or current topics without lingering too long on any one of them, similar to when two strangers meet and start a chit-chat. The chit-chat topics are coronavirus, vacation/travel, and leisure activities. Just as is the case at any point in the conversation, even during the introduction chit-chat the DM ensures that the conversation topic is changed immediately whenever the user expresses the desire to do so. If they do not, Athena ultimately moves on to suggesting one of the topics she can discuss in more depth.

%Initially, we had Athena ask the user for their name at the beginning of the conversation, however, we observed that the accuracy of the ASR was too low for recognizing names, especially non-English ones. We decided that calling a user something that is not quite their name throughout the conversation is highly inappropriate. Alternatively, having them spell their name breaks the flow of the conversation right at the beginning, and that would still not guarantee that the text-to-speech module correctly pronounces  the name. At any rate, people can be very sensitive about their name, and therefore we decided not to take this gamble, and instead we have Athena ask how they are in the first turn.

%\textbf{Nutrition RG.}
%\textcolor{red}{Jurik?}

\section{Analysis and Findings}
\subsection{User Ratings}
\begin{figure*}
    \centering
    \includegraphics[width=0.7\textwidth]{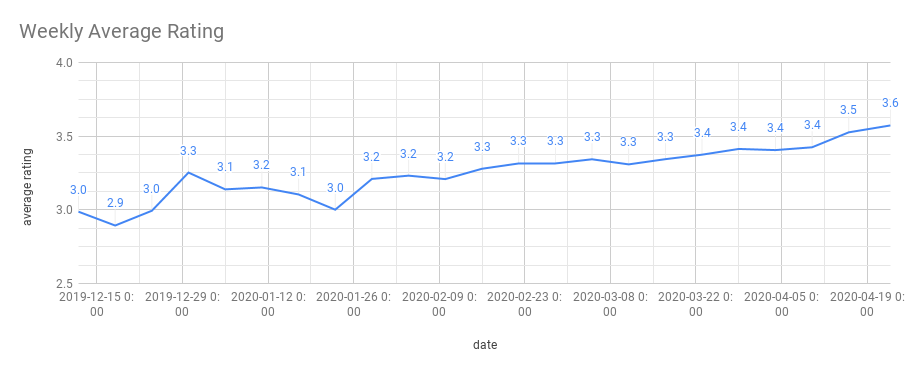}
    \caption{Average user rating of Athena by week.}
    \label{fig:weekly-average-rating}
\end{figure*}

% amazon reviewer action item: Please remove indications of traffic volume
%We collected around 2500 conversations each week since December. 
Figure~\ref{fig:weekly-average-rating} shows the average weekly rating, showing how Athena has steadily improved over this time period. In particular cases we can attribute large ratings jumps to specific innovations. Our low scores in January were due to system engineering issues with timeouts. In mid March, we introduced better topic transition management, resulting in average ratings increase from 3.3 to 3.4. 
On April 10\textsuperscript{th}, we introduced a new introduction flow intended to increase empathy with the user and talk about user interests, leading to an increase from 3.4 to 3.5. On April 20\textsuperscript{th}, we introduced better algorithms for managing interleaving among multiple RGs and tuned transitions between these RGs,  leading to a recent improvement from 3.5 to 3.6.

\begin{figure*}
    \centering
    \includegraphics[width=0.7\textwidth]{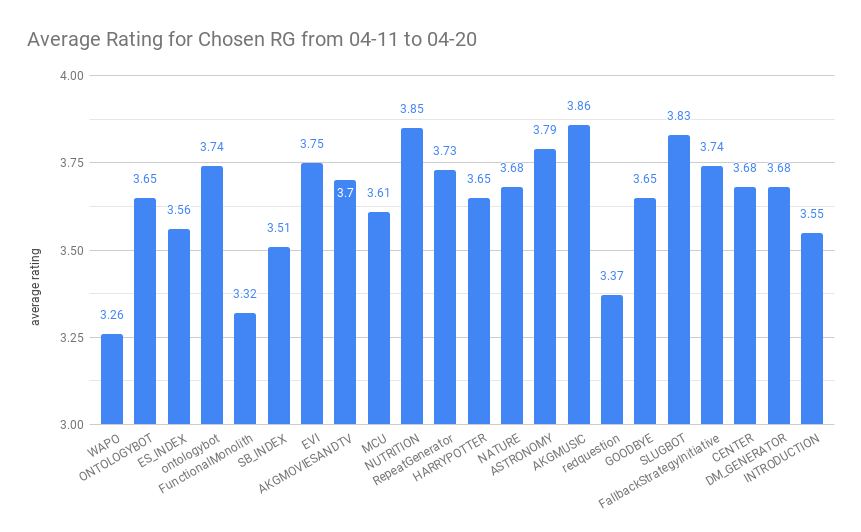}
    \caption{The average rating distribution of Athena's RGs.}
    \label{fig:rg-avg}
\end{figure*}

We also calculate the average rating for each RG. The rating of a conversation  count towards the RG rating whenever that RG is triggered in the conversation. 
Figure~\ref{fig:rg-avg} 
shows the average user rating for the RG chosen by the dialogue manager and its distribution. Some RGs contribute to multiple conversation topics while others contribute to a single topic.
We see that average ratings vary across RGs. The \texttt{WAPO} RG has the lowest average (3.26), but it is also one of our least called upon RGs.
The \texttt{redquestion} RG also has a low average (3.37) which suggests that avoiding inappropriate or controversial topics may negatively effect ratings if the user wishes to talk about them. The \texttt{Nutrition} (3.85), \texttt{AKGMUSIC} (3.86), and \texttt{SLUGBOT} (3.83) RGs have the 3 highest average ratings. 

%\textcolor{red}{We need to add a discussion of the average ratings here.}

\subsection{Experiments with Topical Chat Neural Response Generator}
\label{tc-nrg-sec}

%\begin{wrapfigure}{r}{0.4\textwidth}
%  \centering
%  \includegraphics[width=0.4\textwidth]{figures/TC-NRG-movies-KG-dialogue.png}
%  \caption{Sample Movies  Dialogue with TC-NRG generated system utterances using AKG relations as Knowledge}
%  \label{fig:tc-nrg-movies}
%\end{wrapfigure} 
  
%  \begin{wrapfigure}{r}{0.4\textwidth}
%  \centering
%\includegraphics[width=0.4\textwidth]{figures/TC-NRG-movies-KG-dialogue.png}
% \includegraphics[width=0.4\textwidth]{figures/delay-without-NRG.png}
% \includegraphics[width=0.4\textwidth]{figures/delay-w-nrg.png}
%  \caption{Athena's response latency distribution without and with TC-NRG.}
%   \label{fig:response-delay}
%\end{wrapfigure} 

\begin{figure}[h]
    \centering
    \begin{subfigure}[b]{\linewidth}
        \includegraphics[width=\linewidth]{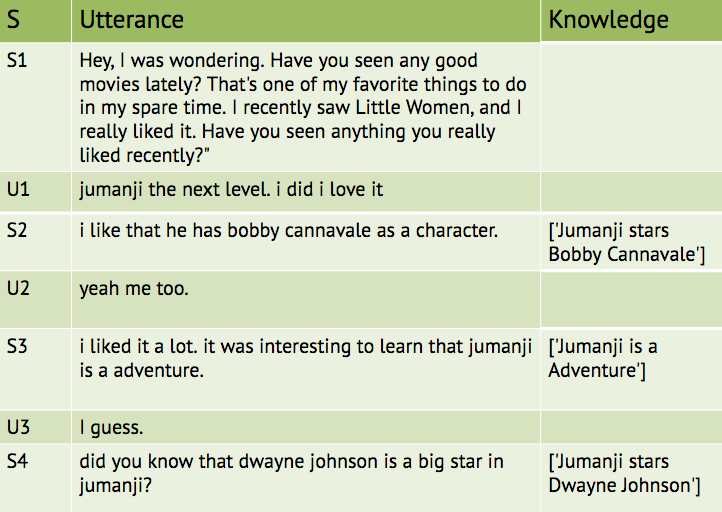}
        \caption{Sample movies  dialogue with TC-NRG generated system utterances using AKG relations as knowledge.}
    \end{subfigure}
    \hfill
    
    \begin{subfigure}[b]{\linewidth}
        \includegraphics[width=\linewidth]{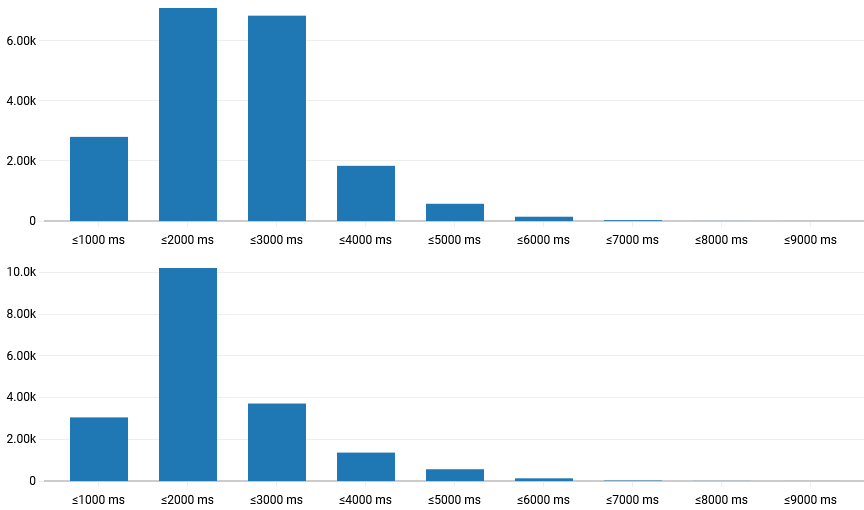}
        \caption{Response latency distribution with/without TC-NRG over a 24-hour period of time (April~22\textsuperscript{nd} and April~15\textsuperscript{th}, respectively).}
    \end{subfigure}
    \caption{TC-NRG example dialogue, and the module's effect on Athena's response latency.}
    \label{fig:tc-nrg-example-delay}
\end{figure}

We integrated Amazon's Topical Chat Neural Response Generator (TC-NRG) into Athena and tested it both in the running system and off-line, for two kinds of knowledge -- Wikipedia and AKG facts. We note that it was unclear how  TC-NRG would perform: the dialogue context for Athena is different than the TC corpus, in particular utterances in the TC corpus are about 20 words long, while user utterances in Athena  are typically shorter \cite{gopalakrishnan2019topical}. Furthermore performance in Athena is dependent on selecting relevant knowledge to provide to TC-NRG, while the TC training corpus provides "oracle" relevant knowledge. We collected a sample of 102 generated utterances using 5 turns of context
with Wikipedia as a knowledge source, and 54 generated utterances using AKG. A sample dialogue with  generated utterances using AKG is in Figure~\ref{fig:tc-nrg-example-delay}a, and sample utterances for Wikipedia are in Table~\ref{table:TC-NRG-outputs}. We logged all TC-NRG responses and then carried out a human evaluation to determine the feasibility of deploying TC-NRG in Athena as a full-fledged response generator.

Our first observation is that, when deployed in the live system, TC-NRG significantly increased Athena's response delay (see Section~\ref{subsec:response_latency}). During the window of time tested, there was an issue that led to occasional timeouts of the NRG service. This issue was later resolved, but we had completed our evaluation at that point. Athena's normal response latency distribution is shown in the bottom part of Figure~\ref{fig:tc-nrg-example-delay}b, while latencies with the TC-NRG enabled are in the top part. We thus only ran TC-NRG in the production system for 6 days, and carried out our other experiments off-line.

\begin{table*}[t]
       \centering
       \begin{tabular}{l c c c c c c c}
            \toprule
            \textbf{Knowledge} & \textbf{Und} & \textbf{Nat} & \textbf{MCon} & \textbf{Int} & \textbf{UK} & \textbf{OQ}  \\
            \midrule
            \textbf{Wikipedia} &  0.66 & 1.90 & 2.11 & 1.49 & 0.42 & 2.13 \\
            \textbf{AKG Movie Facts} & 1 & 2.57 & 2.75 & 2.57 & 0.77 & 3.40 \\
            \bottomrule\\
       \end{tabular}
       \caption{USR ratings by knowledge source.}
       \label{table:usr-ratings}
\end{table*}

To evaluate quality, we labelled the 156 TC-NRG output responses using the 6 utterance quality metrics that form the basis of the USR metric \cite{mehri-USR20}: {\it Understandable} (Und; 0-1), {\it Natural} (Nat;1-3), {\it Maintains Context } (MC; 1-3), {\it Interesting} (Int; 1-3), {\it Uses Knowledge} (UK; 0-1) and {\it Overall Quality} (OQ; 1-5). The metrics for each knowledge source are  in Table~\ref{table:usr-ratings}, showing which that TC-NRG does much better with knowledge realized via templates from AKG, than it does with Wikipedia entries.  The templated KG relations tend to be simple sentences, and are utilized much more frequently in the utterances ({\it Uses Knowledge} is 0.77 for AKG vs. 0.42 for Wikipedia). We cannot ascertain whether this is due to abundant training data for movies as opposed to arbitrary Wikipedia topics, or the simple structure of the AKG templated utterances.
Note that  utterances generated using  AKG knowledge are better across the board for all metrics: they are more understandable, natural, make better use of context, are more interesting and have higher overall quality.  On the basis of this comparison, we believe we could possibly use TC-NRG as an additional RG for the  movies, music and sports domains where we already access and use KG relations. We would first have to further experiment with sports and music to test whether the findings for movies holds there, and we would also have to address the response latency issues.
 
\begin{table*}
    \centering
    \begin{tabular}{p{1.2in} p{2.3in} p{1.5in}}
        \toprule
        Output & Context & Knowledge  \\
        \midrule %\\
        i like the lakers too. they are the best basketball team.  & ...,
        % \{'agent': 'User', 'message': 'wow'\}, \{'agent': 'Bot', 'message': 'I found an interesting piece of trivia on the book Frankenstein. The Monster in Frankenstein has no name, but Mary Shelley once referred to him as Adam.'\}, \{'agent': 'User', 'message': "really that's weird"\},
        \textbf{BOT:} I'm also really interested in sports. I like to watch the games on the TV. Recently, I've been watching a lot of basketball. Are there any teams that you follow? \textbf{USER:} i like the lakers 
        & The Los Angeles Lakers are an American professional basketball team based in Los Angeles.....
        %The Lakers compete in the National Basketball Association , as a member of the league's Western Conference in the Pacific Division.}
        \\ \hline
        i have been to the isle royale national park in michigan. i used to play it all the time when i was a kid. i guess the game is called "game of thrones" or something
        & ..., \textbf{USER:} let's talk about games. \textbf{BOT:} There are some really beautiful national parks in the U.S.! Have you been to the Isle Royale National Park by any chance? It's in Michigan. \textbf{USER:} alexa let's talk about video games
          & A video game is an electronic game that involves interaction with a user interface to generate visual feedback on a two- or three-dimensional video display ....
          %device such as a touchscreen, virtual reality headset or monitor/TV set.} 
          \\ \hline
          that sounds like a great place to meet, i think it's a great place to meet, i'll have to check it out. & ...,
         %\{'agent': 'User', 'message': "alexa let's talk about video games"\}, 
         %\{'agent': 'Bot', 'message': "So, what do you think, Would you rather play a game that's innovative, but clunky, or a game that's unoriginal, but smooth?"\}, \{'agent': 'User', 'message': "i don't riginal but smooth"\}, 
        \textbf{BOT:} Right. I would probably choose an unoriginal game but smooth too. I'd rather play a game that's unoriginal but smooth. I don't like struggling with a game as I'm playing it, even if the pay-off is extremely good. \textbf{USER:} how about we talk about five nights at freddy's 
        & Five Nights at Freddy's  is a media franchise based on an indie video game series created, designed, developed, and published by Scott Cawthon ....
        %for Microsoft Windows, iOS, and Android. The series is centered on the fictional Freddy Fazbear's Pizza, a pastiche of pizza restaurants such as Chuck E. Cheese's and ShowBiz Pizza Place.}
        \\
        \bottomrule
        \\
    \end{tabular}
    \caption{Sample Outputs of Topical Chats NRG with Wikipedia Knowledge}
    \label{table:TC-NRG-outputs}
\end{table*}

We then conducted a further qualitative analysis of the outputs. On the positive side, Figure~\ref{fig:tc-nrg-example-delay} shows that TC-NRG seems to successfully integrate knowledge into the context in a natural way, serving as a stylistic paraphraser of the original knowledge by using phrases such as {\it it was interesting to learn} and {\it did you know that}. It also appears to generate pronouns and other anaphora appropriately,  However, 
TC-NRG also displays limitations similar to other neural generators. For example,  the TC-NRG tends to respond "I like X" whenever the user says   "I like X". See Table~\ref{table:TC-NRG-outputs} Row 1, and S3 and S4 in Figure~\ref{fig:tc-nrg-example-delay}. And despite its knowledge-grounding, in some cases, it combines information inconsistently, as in Row 2 of Table~\ref{table:TC-NRG-outputs}, where it reuses information from the context instead of constraining itself to talk about the provided knowledge on video games, while in Row 3, it simply ignores knowledge that "Five Nights at Freddies" is a video game. However, improvements are consistently being made in knowledge-grounded dialogue generation, and we are currently experimenting further with controllable generation for knowledge-grounded dialogue \cite{hedayatnia2020policy}.

\subsection{Effects of Response Latency on Ratings}
\label{subsec:response_latency}

\begin{figure*}[h]
    \centering
    \includegraphics[width=0.74\textwidth]{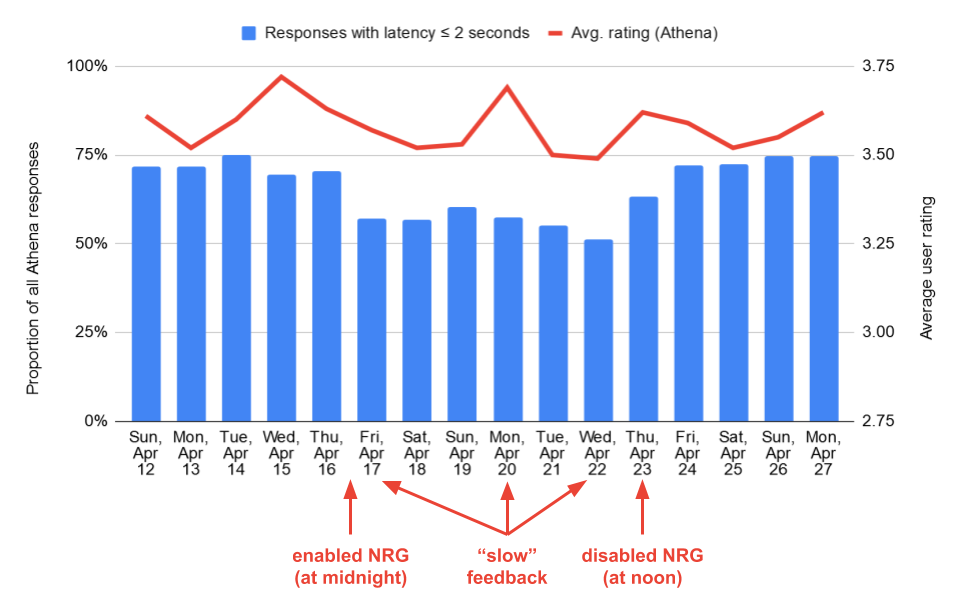}
    \caption{Athena's average user ratings and response latencies before, during, and after the week the NRG module was enabled for.}
    \label{fig:response-latency-vs-rating}

\end{figure*}

Besides observing a significant increase in Athena's average response latency immediately after deploying the NRG module, starting around the same time, we also noticed a drop in Athena's user ratings that lasted for several days. Figure~\ref{fig:response-latency-vs-rating} shows Athena's ratings along with response latencies over the period of two weeks around the time when the NRG module was enabled in the system. Note that the module was not contributing to the response pool at this point, and had thus no effect on Athena's response content or quality while enabled.

Initially, we attributed Athena's lower post-deployment ratings to the usual dip in average ratings on the weekends (Friday through Sunday). Nevertheless, when analyzing the user feedback, we noticed that we had started receiving comments about Athena's responses being slow, which had not happened a single time over the period of 4 weeks before that.\footnote{Transcriptions of user feedback were delivered by the competition organizer with an approximately 5-day delay, and there were typically less than 15 users per day that provided feedback for Athena.} 

As can be seen in the figure, there is a clear downward trend in Athena's ratings between April 17\textsuperscript{th} and 22\textsuperscript{nd}.\footnote{We consider the sudden peak on April 20\textsuperscript{th} an outlier, which is something we sporadically observed in Athena's ratings, suggesting the user ratings are overall rather noisy.} Athena's ratings then immediately picked up on April 23\textsuperscript{rd} when we disabled the NRG module. We did not observe a similar trend in user ratings across all socialbots during the same time period. For the next couple of days, Athena then followed the expected overall downward trend at the beginning of the weekend.

While we cannot prove that it was indeed the increased response latency that caused Athena's ratings to drop, since our other experiments showed the user ratings to be very noisy, we believe these results show that users are sensitive to the socialbot's response latency. The balance here seems rather fragile, with additional 300-400 milliseconds appearing to change users' perception of the system's responsiveness.
% While this observation led us to disabling the NRG module until we further optimize our system to accommodate such an increase in latency without a negative impact on the ratings, it also reinforced the importance of using progressive responses, such as Athena does for grounding, in order to minimize the perceived response latency of the socialbot.
This observation reinforces the importance of using progressive responses, such as Athena does for grounding, in order to minimize the perceived response latency of the socialbot.

The observed user sensitivity to Athena's response latency was also the reason why we ultimately abandoned the use of our neural data-to-text generator~\cite{juraska2018deep}. Trained on the ViGGO dataset~\cite{juraska2019viggo}, it was restricted to a single topic, but it increased the latency more significantly than TC-NRG. The ViGGO dataset was still instrumental in the development of Athena's video game RG, as it was designed around the dialogue acts defined therein.

% \begin{figure}
%   \centering
%     \subfloat[average rating]{\label{fig:rg-avg-rating}\includegraphics[width=0.7\textwidth]
%   {figures/analysis_avg_rating_chosen_rg.png}}
%   \subfloat[distribution]{\label{fig:rg-distribution}\includegraphics[width=0.35\textwidth]
%   {figures/analysis_distribution_chosen_rg.png}}
  
%  \label{fig:rg_analysis}\caption{The average rating and distribution of the RG.}
% \end{figure}

% \begin{figure}
%   \centering
%     \subfloat[average rating]{\label{fig:ctk-avg-rating}\includegraphics[width=0.7\textwidth]
%   {figures/analysis_avg_rating_ctk_topic.png}}
%   \subfloat[distribution]{\label{fig:ctk-distribution}\includegraphics[width=0.35\textwidth]
%   {figures/analysis_distribution_ctk_topic.png}}
  
%  \label{fig:ctk_topic_anlysis}\caption{The average rating and distribution of the ctk\_topic.}
% \end{figure}

\subsection{User Testing: Addressing Topic Coverage and Individual User Characteristics}

We conducted several user studies in order to understand better how individual user characteristics affect system ratings, and to collect qualitative feedback on design decisions for Athena's dialogue manager.  We recruited seven Psychology and Linguistics undergraduates to conduct two user evaluations involving 86 users who each used the system for 20-40 minutes. Our focus was on system coverage, user characteristics and user expectations - key design issues identified in prior evaluations \cite{bowden2019cui}. We also collected general user feedback about conversational flows and bugs.

\noindent{\bf Study 1: Users Prefer to Select Topics and Are Influenced by Prior Expectations.} One experimental variable was Topic Coverage.  Allowing users to choose their own conversational topics increases the likelihood that they will engage with that topic, but may result in them selecting topics the system knows little about. Our first study therefore compared user reactions to system-directed versus user-driven topics. We expected higher user evaluations for system-directed interactions as Athena can talk more fluently about these topics, even though users may be less interested in these.

A second experimental variable was User Characteristics and Expectations. Prior research shows that user expectations are important in determining reactions to new technologies \cite{paepcke2010judging}. Users with little conversational system experience are more likely to rate technology negatively after using it because their unrealistic expectations are not met. The study therefore constructed user profiles using personality metrics and documented users’ experiences with, and expectations about, conversational technologies. We also gathered informal user reactions to the system, including what topics users would like to discuss with the system. 

We first profiled 32 users assessing their personality, system experience, and expectations. Users then conversed with Athena twice, once when they chose their own topics (user-topics condition) and a second time when they were told to choose from the following topics (system-topics condition) for which Athena has excellent coverage: Dinosaurs, Animals, Astronomy, Sports, Movies, Music, Nutrition, Books. Conversations lasted between 7 and 22 mins. After each conversation, users rated their interaction. 

Results showed, contrary to our expectations, that users rated conversations more highly in the user-topics condition, when they were allowed to select their own topics (p$=$0.02). There were also large individual differences in reactions based on personality: extraverts (p$=$0.019) and more conscientious users (p$=$0.003) rated the system more highly overall. We also found that people with higher initial expectations rated the system lower after usage (p$=$0.015), perhaps because they had little prior experience with speech systems. However qualitative comments indicated no clear consensus about topics that users would like to see covered. 

\noindent{\bf Study 2: Improved Ratings for System Topics and Reduced Individual Differences.}
These results suggested a design challenge. Even though users preferred to choose their own topics, we decided against trying to broaden overall coverage as there was no consensus about which new topics should be added. Instead we chose to enhance coverage of our existing topics, to make them more engaging. By making these topics more robust we hoped to improve perceptions for those with little prior experience (and high expectations) of speech systems. 

A second round of user testing evaluated an improved version of the system, using the same method with 54 participants. The results were encouraging. Overall interaction ratings had improved over the previously tested version (p$=$0.046). This was attributable to improved ratings for system-topic conversations (p$=$0.04) while user-topic conversation ratings were unchanged (p$=$.99). Further, even though the actual topics were unchanged from the prior study, participants’ perceptions of their control over the system had increased by 51\% (p$=$0.0001). These observations were reflected in user comments. More importantly we reduced individual differences; both personality and expectation effects had disappeared, suggesting the system was more robust to individual differences and prior experience with conversational systems. 

Overall the two studies showed how we were able to incorporate user-centric methods to address and test a key coverage problem with our system. This feedback meant we were able to boost overall system ratings on system topics by 17\%, as well as making the system more broadly resilient to user demographics and expectations. 

%\subsubsection{what Jurik just did with the free form user responses?}

\section{Conclusion}
Here we describe a conversational agent for spoken dialogue named Athena,
which competed as an Alexa Prize Socialbot in 2019/2020. 
Athena's dialogue dialogue manager sends response requests to collections of response generators through use of discourse constraints. This approach allows the dialogue system to use dynamic discourse planning. 
Furthermore, Athena is able to incorporate responses from a wide variety of sources, such as real-time knowledge graph walks, as well as a modular conversation flow framework. We develop a new named entity resolution system that incorporates a large knowledge base of entities as well as an ensemble of publicly available named entity linking systems. We analyze a sample of Athena's conversations collected during the semi-finals phase of the Alexa Prize 2019. We report on several user studies that show users prefer to select conversation topics and user's prior expectations influence conversation quality ratings.

%\subsubsection*{Acknowledgments}
%Use unnumbered third level headings for the acknowledgments. All
%acknowledgments go at the end of the paper. Do not include
%acknowledgments in the anonymized submission, only in the final paper.

%\section*{References}
%References follow the acknowledgments. Use unnumbered first-level
%heading for the references. Any choice of citation style is acceptable
%as long as you are consistent. It is permissible to reduce the font
%size to \verb+small+ (9 point) when listing the references. {\bf
%  Remember that you can use a ninth page as long as it contains
%  \emph{only} cited references.}

%\bibliography{athena,nl}
\bibliography{nl}
\bibliographystyle{acl_natbib}

\end{document}